# FusionNet: A deep fully residual convolutional neural network for image segmentation in connectomics

Tran Minh Quan[1], David G. C. Hildebrand[2] and Won-Ki Jeong[1*]
[1]Ulsan National Institute of Science and Technology (UNIST),
[2]Harvard University

`quantm@unist.ac.kr, david@hildebrand.name, wkjeong@unist.ac.kr`

## Abstract

*Electron microscopic connectomics is an ambitious research direction with the goal of studying comprehensive brain connectivity maps by using high-throughput, nanoscale microscopy. One of the main challenges in connectomics research is developing scalable image analysis algorithms that require minimal user intervention. Recently, deep learning has drawn much attention in computer vision because of its exceptional performance in image classification tasks. For this reason, its application to connectomic analyses holds great promise, as well. In this paper, we introduce a novel deep neural network architecture, FusionNet, for the automatic segmentation of neuronal structures in connectomics data. FusionNet leverages the latest advances in machine learning, such as semantic segmentation and residual neural networks, with the novel introduction of summation-based skip connections to allow a much deeper network architecture for a more accurate segmentation. We demonstrate the performance of the pro- posed method by comparing it with state-of-the-art electron microscopy (EM) segmentation methods from the ISBI EM segmentation challenge. We also show the segmentation results on two different tasks including cell membrane and cell body segmentation and a statistical analysis of cell morphology.*

## 1. Introduction

"How does the brain work?" This question has baffled biologists for centuries. The brain is considered the most complex organ in the human body, which has limited our understanding of how relating its structure is related to its function even after decades of research [21]. Connectomics research seeks to disentangle the complicated neuronal circuits embedded within the brain. This field has gained substantial attention recently thanks to the advent of new serial-section electron microscopy (EM) technolo-gies such as the automated tape-collecting ultramicrotome (ATUM) [13] (see Figure 1 for an example of EM image and its cell membrane segmentation). The resolution afforded by EM is sufficient for resolving tiny but important neuronal structures that are densely packed together, such as dendritic spine necks and synapses. These structures can be as small as only tens of nanometers in width [15]. Such high-resolution imaging results in the generation of enormous datasets, approaching one petabyte for only a relatively small tissue volume of one cubic millimeter. Therefore, handling and analyzing the resulting datasets is one of the most challenging problems in connectomics.

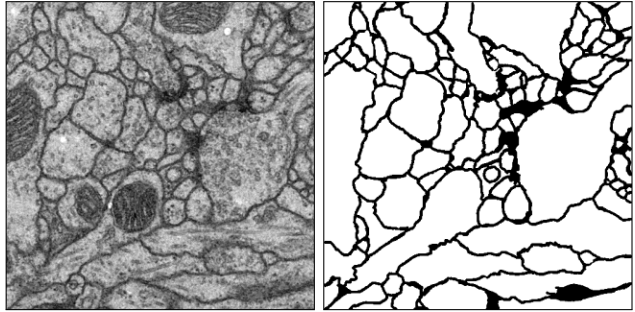

Figure 1. An example of EM image (left) and its cell membrane segmentation result (right).

Early connectomics research focused on the sparse reconstruction of neuronal circuits [4, 5], i.e., tracing only a subset of neurons in the data by using manual or semi-automatic tools [7, 17, 29]. Unfortunately, this approach requires too much human interaction to scale well over the vast amount of EM data that can be collected with technologies such as ATUM. Because of this, the field has been limited in the number of datasets that have been thoroughly annotated and analyzed. In addition, multi-scale reconstruction, including dense reconstruction in the region of interest, has gained popularity recently because it can reveal low-level structural information that is not available in sparse reconstruction or functional imaging [18]. Therefore, developing scalable and automatic image analysis algorithms is an important and active research direction in the field of connectomics.

Although some EM image processing pipelines (e.g., RhoANA [18]) use conventional, light-weight pixel classifiers, the majority of recent automatic image segmentations for connectomics rely on deep learning. Earlier automatic segmentation work using deep learning has mainly focused on patch-based pixel-wise classification based on a convolutional neural network (CNN) for affinity map generation [32] and cell membrane probability estimation [10]. However, one limitation of applying a conventional CNN to EM image segmentation is that per-pixel network deployment can be highly expensive in consideration of the tera- to peta-scale EM data size. For this reason, a more efficient, scalable deep neural network will be important for image segmentation of the large datasets that can now be produced [8, 25]. The main idea behind these approaches is to extend a fully convolutional neural network (FCN) [23], which uses encoding and decoding phases similar to an autoencoder for the end-to-end semantic segmentation problem.

The motivation of the proposed work stems from our recent research effort to develop a deeper neural network for end-to-end cell segmentation with higher accuracy. We observed that the current state-of-the-art deep neural network for end-to-end segmentation (i.e., U-net) [25] shares a similar limitation with the conventional CNN in increasing network depth, as discussed in residual CNN [14]. To address this problem, we propose a novel extension of U-net by using residual layers in each level of the network and introducing summation-based skip connections to make the entire network much deeper than U-net. Our segmentation method produces an accurate result that is competitive with other state-of-the-art EM segmentation methods without complicated post-processing enhancements. The main contribution of this study can be summarized as follows:

- We introduce a novel end-to-end automatic EM image segmentation method using deep learning. The proposed method is based on a variant of U-net and residual CNN with a novel summation-based skip connections that make the proposed architecture a *fully residual deep CNN*. Therefore, it directly employs residual properties within and across levels and hence grants the possibility of building a *deeper* network with higher accuracy.

- We demonstrate the performance of the proposed deep learning architecture by comparing it with the state-of-the-art EM segmentation methods listed in the leader board of the ISBI 2012 EM segmentation challenge. Our method outperformed most of the top-ranked methods in terms of segmentation accuracy without complicated post-processing enhancement.

- We introduce a *data enrichment* method specifically built for EM data by collecting all the orientation variants of the input images (eight in the 2D case for the combination of flipping and rotation). We used the same data duplication process for deployment: the final output can be a combination of 8 different probability values, which can increase the accuracy of the method.

- We demonstrate the flexibility of the proposed method on two different EM segmentation tasks; one is cell membrane segmentation on a Drosophila EM dataset, and the other is cell nucleus segmentation on a whole-brain larval zebrafish EM dataset.

## 2. Related work

In the last five years, deep learning [20] has gained much attention, largely because it has surpassed the human level in solving many complex problems. It is comprised of many perceptron layers that form a deep neural network. In visual recognition tasks, this type of architecture can learn to recognize patterns such as handwritten digits and other features of interests [19] in images hierarchically [35]. However, the main drawback of using deep neural network is that it requires a huge amount of data for training the network. In order to overcome this issue, researchers have started to collect a large database [26] which contains millions of images from hundreds of categories. Since then, many advanced architectures have been introduced including VGG [28], Googlenet [31]. Computers are now able to mimic artistic painting to produce new pictures by transferring the style from one image to another [12]. In addition, researchers are also actively working on extending deep learning methods for medical image data beyond the scope of natural images [9]. These approaches impose vast changes in automatic classification and segmentation on other image modalities, such as CT [36] and MRI [16]. These studies have opened a revolutionary era in which software can self-program to achieve or even outperform human capabilities in image processing and computer vision areas. Deep learning has been quickly adopted by connectomics research for automatic EM image segmentation. One of the earliest applications to EM segmentation was made by Ciresan et al. [10]. This method involves the straightforward application of a CNN for pixel-wise membrane probability estimation and it won the ISBI 2012 challenge [2]. As new deep learning methods are introduced, automatic EM segmentation techniques evolves, as well. One notable recent advancement in the machine learning domain is the introduction of a fully convolutional neural network (FCN) [23] for the end-to-end semantic segmentation problem. Inspired by this work, many successive variants of FCN have been proposed for EM image segmentation. Chen et al. [8] proposed multi-level upscaling layers and their combination for final segmentation. A new-post processing step, namely lifted multi-cut [3], was also introduced to refine the segmentation. Ronneberger et al. [25] presented skip connections for concatenating feature maps in their U-net architecture. Although U-net and its variants can learn multi-contextual information from the input data, they are limited in the depth of the network they can construct because of the vanishing gradient problem. Recently, the 3D extension of U-net was proposed for confocal microscopy segmentation [9]. In the image classification task, on the other hand, shortcut connections and direction summations [14] allow gradients to flow across multiple layers during the training phase. Overall, these related studies inspired us to propose a fully residual convolutional neural network for analyzing connectomic data. Work that leverages recurrent neural network (RNN) architectures can also accomplish this segmentation task [30]. In fact, the membrane-type segmentation approach is a crucial step for connected component labeling to resolve false splits and merges during the post-processing of probability maps [11, 24].

## 3. Method

### 3.1. Network architecture

Figure 2 is a pictorial description of the proposed network architecture. Our network is built based on the architecture of a convolutional autoencoder, which consists of an encoding path (upper half of the network, from 640×640 to 40×40) to retrieve the features of interest and a symmetric decoding path (lower half of the network, from 40×40 to 640×640) that enables the prediction of synthesis. Each encoding/decoding path consists of multiple levels, i.e., resolutions, to extract features in different scales. Four types of basic building blocks are used to construct the proposed network. Each *green block* is a regular convolutional layer followed by rectified linear unit activation and batch normalization (omitted in the figure for simplicity). Each *violet block* is a residual layer that consists of three convolutional blocks and a residual skip connection. A maxpooling layer is located between levels in the encoding path to perform downsampling for feature compression (*blue blocks*). The deconvolutional layer (*red blocks*) is located between levels in the decoding path to up-sample the input data using learnable interpolations. A more detailed specification of our network, such as the number of feature maps and their sizes, is provided in Table 1.

One major difference between our network and U-net is the skip connection. Each step in the decoding path begins with a deconvolutional (red) block that un-pools the feature from a coarser level (i.e., resolution), and then it is merged using a pixel-wise *addition* with the feature map from the same level in the encoding path by a *long* skip connection (Figure 3 right). In addition, there is a *short* skip connection in a residual block (the violet blocks in Figure 3 right) that allows for a direct connection from the previous layer in the same encoding/decoding path. Unlike our method, U-net uses the concatenation of feature maps via only the long skip connection. By replacing a concatenation with an addition, our network becomes a *fully* residual network and some issues in deep network (i.e., gradient vanishing) can be handled effectively. In addition, our network has nested short and long skip connections that help information flow within and across levels in the network.

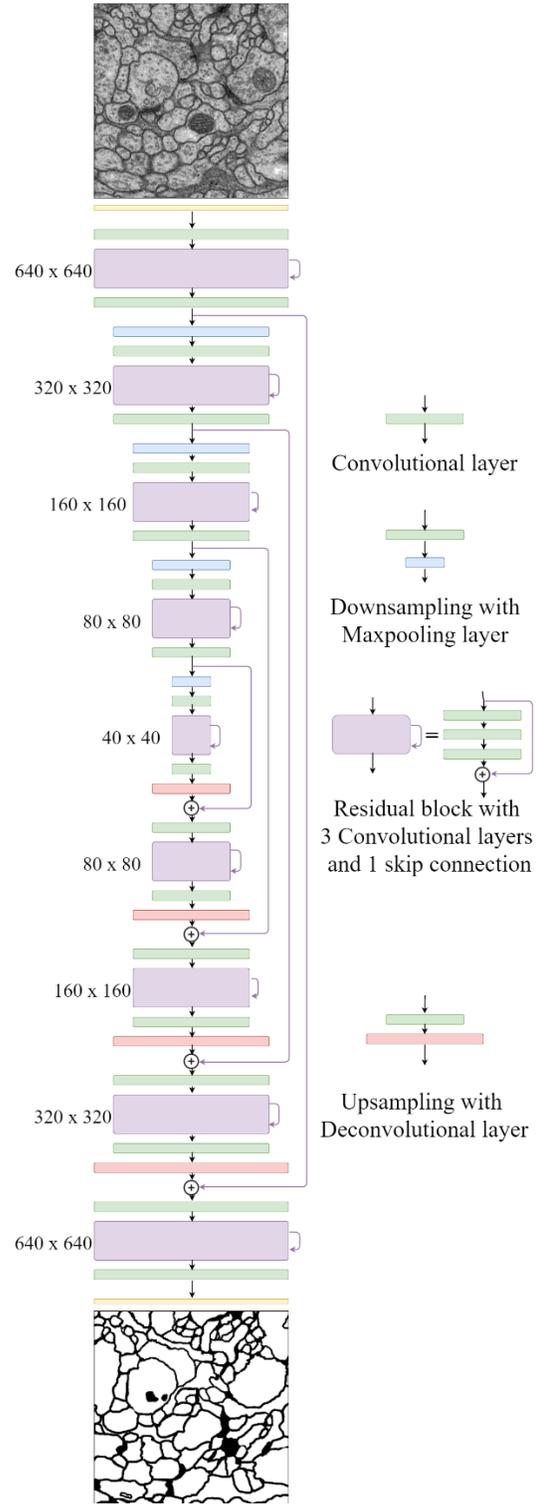

Figure 2. The architecture of the proposed network. An illustration of the encoding path (top to middle) and the decoding path (middle to bottom). Each intermediate residual block contains a residual skip connection within the same path, but the nested residual skip connections connect two different paths.

Table 1. Architecture of the proposed network

| Block type | Ingredients | Size of feature maps |
|---|---|---|
| inputs | | 640×640×1 |
| down 1 | conv + res + conv | 640×640×64 |
| | + maxpooling | 320×320×64 |
| down 2 | conv + res + conv | 320×320×128 |
| | + maxpooling | 160×160×128 |
| down 3 | conv + res + conv | 160×160×256 |
| | + maxpooling | 80×80×256 |
| down 4 | conv + res + conv | 80×80×512 |
| | + maxpooling | 40×40×512 |
| bridge | conv + res + conv | 40×40×1024 |
| upscaling 4 | deconv + merge + | 80×80×512 |
| | conv + res + conv | 80×80×512 |
| uspcaling 3 | deconv + merge + | 160×160×256 |
| | conv + res + conv | 160×160×256 |
| upscaling 2 | deconv + merge + | 320×320×128 |
| | conv + res + conv | 320×320×128 |
| upscaling 1 | deconv + merge + | 640×640×64 |
| | conv + res + conv | 640×640×64 |
| output | conv | 640×640×1 |

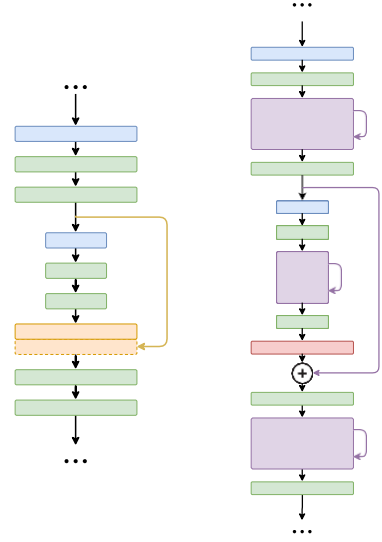

Figure 3. Difference between core connections of U-net [25] (left) and the propose network (right). Note that our network is not only much deeper compared to U-net but also a fully residual network due to summation-based skip connections.

In the encoding path, the number of feature maps is doubled whenever downsampling is performed. After passing through the encoding path, the bridge level (i.e., 40×40 layer) also has a residual block and starts to expand the feature maps in the following decoding path. In the decoding part, the number of feature maps is halved per level to maintain the network symmetry. Note that each residual block has two convolutional layers, before and after the block. These convolutional layers serve as a connector to bridge the input feature maps and the residual block because the number of feature maps from the previous layer may differ from that of the residual block. Another benefit of having a convolutional layer on each side of the residual block is that the entire network becomes perfectly symmetric (see Figure 2).

The proposed network performs an end-to-end segmentation from the input EM data to the final prediction of the segmentation. We train the network with pairs of images (the EM image and its corresponding label image), compare the output with manual segmentation, and use its mean-square-error loss function to back-propagate to adjust the weights of the network. Once the network is trained appropriately (i.e., its loss values are relatively small), it is ready to be deployed on the test data.

### 3.2. Data augmentation

**Data enrichment:** Different EM images typically share similar orientation-independent textures in micro-structures such as mitochondria, axons, synapses, etc. We therefore can enrich our data by forming an additional seven other sets of raw images and their labels. Figure 4 shows the variants of eight orientations when we perform image enrichment. The letter 'g' in the middle represents a simpler view of the generated direction. For this permutation, we rotate the dataset by 90°, 180°, and 270°, respectively and we take the reflections of those four sets to form the entire training data. Note that because the images and labels have been enriched eight times, other on-the-fly data augmentation techniques such as random rotation, flipping, or transposition can be turned off with the exception of elastic deformation.

**Elastic field deformation:** To avoid the overfitting case (i.e., network remembers the training data), we perform im-age elastic deformation on the entire enriched dataset per every training epoch. This strategy is common in machine learning, especially for deep networks, to overcome the limitation of small training dataset sizes. Figure 5 is a pictorial description of this procedure. We first initialize a random sparse vector field (12×12) whose amplitudes at the boundaries have vanished (zero amplitude). This field is then interpolated to the size of the original images and used to warp both image pairs (raw data and the corresponding labels) to form new ones. The flow map is randomly generated so that the image can be deformed differently per epoch.

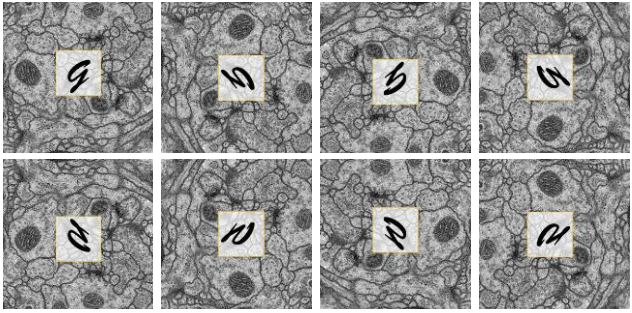

Figure 4. Eight variations of image orientations. By doing this, the input data size is increased by eight times.

**Random noise, boundary extension, random shuffle and cross-validation:** During the training phase, we randomly add Gaussian noise to the raw image (mean $\mu = 0$, variance $\sigma = 0.1$). The dataset is then padded with the mirror reflections of itself across the boundary (*radius* = 64*px*) to grant the statistical neighboring description of the pixels that are near the borders. That is why our network accepts the image size 640×640, which is 128*px* larger than the original size. On the other hand, the convolution mode we used is "same", which leads the final segmentation to have an identical size to the input image. For the final result, we simply crop the prediction to eliminate unnecessary padding regions. To this end, before fetching the enriched data for the model, we also shuffle the order of the dataset and perform a three-fold cross validation to improve the generalization of our method.

### 3.3. Experimental Setup

The proposed deep network is implemented using Keras open-source deep learning library [1]. This library provides an easy-to-use high-level programming API written in Python, and Theano or TensorFlow can be chosen for a backend deep learning engine. Training and deployment of the network is conducted on a PC equipped with an Intel i7 CPU with a 32 GB main memory and an NVIDIA GTX Geforce 1080 GPU. Since we use the data enrichment method that duplicates the input image by applying rota-

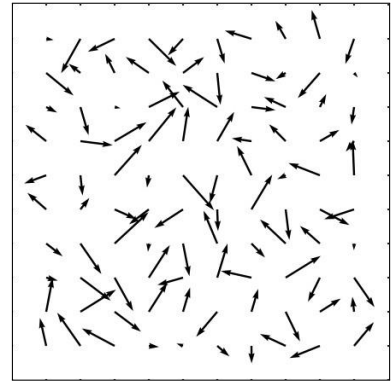

(a) A random sparse vector flow

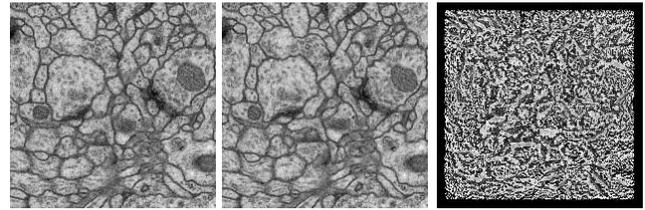

(b) Warping on the image: before, after, difference.

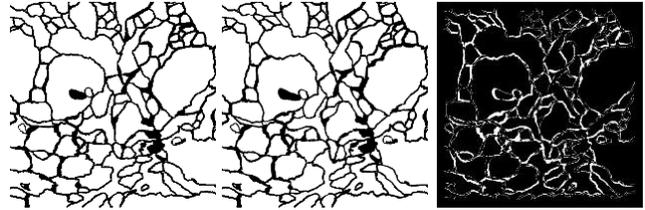

(c) Warping on the label: before, after, difference.

Figure 5. Elastic field deformation example.

tion and mirroring transformations for training, we apply the same data transformation for deployment and average the results.

## 4. Results
### 4.1. Drosophila EM data

The Drosophila ventral nerve cord serial-section electron microscopy data analyzed here was captured from a first instar larva and was previously reported [6]. Both a training and a test dataset were provided as part of the ISBI two-dimensional electron microscopy segmentation challenge [2]. Each was acquired at an anisotropic 4×4×50 nm$^3$ vx$^{-1}$ resolution with transmission EM. The datasets were chosen in part because they contained some noise and small image alignment errors that frequently occur in serial-section EM datasets. For training, the provided set included a 2×2×1.5 um$^3$ volume imaged from 30 sections and publicly available manual segmentations. For testing, the provided set included only image data, with segmentations kept

Table 2. Evaluation on drosophila EM dataset.

| Methods | $V_{rand}$ | $V_{info}$ |
|---|---|---|
| Our approach (Nov 2016) | **0.978042575** | **0.989945379** |
| IAL IC [22] | 0.977345721 | 0.989240736 |
| Masters [34] | 0.977141154 | 0.987534429 |
| CUMedVision [8] | 0.976824580 | 0.988645822 |
| LSTM [30] | 0.975366444 | 0.987425430 |
| Our approach (May 2016) | 0.972797555 | 0.987597732 |
| U-net [25] | 0.972760748 | 0.986616590 |

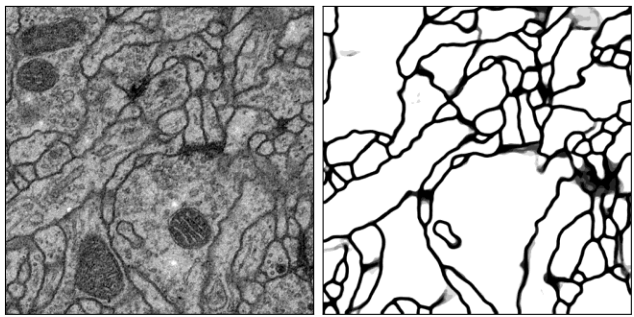

Figure 6. Prediction on test data (Slice 28/30).

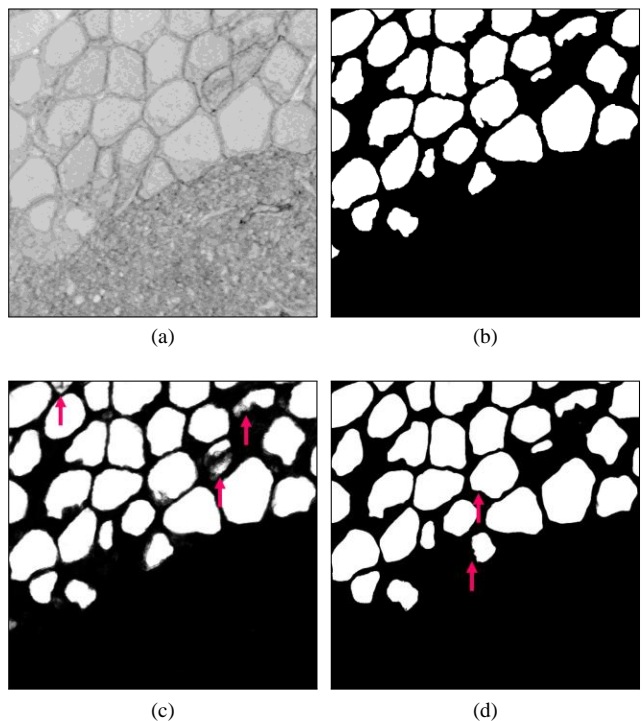

Figure 7. Top row: Raw image (a) of Slice 195/512 on the test set and its manual segmentation (b); Bottom row: U-net [25] result (c) and our result (d). Note that our result shows less false-positive and more accurate segmentation (see red arrows).

Table 3. Evaluation on zebrafish EM dataset.

| Methods | Our approach | U-net [25] |
|---|---|---|
| $V_{rand}$ | 0.991844302 | 0.987366177 |
| $V_{info}$ | 0.994208722 | 0.992482059 |
| $V_{dice}$ | 0.946099985 | 0.908491647 |

private for the assessment of segmentation accuracy [2].

Figure 6 illustrates the results of our segmentation of the test data without any sophisticated post processing step (Slice 28/30). As depicted in this figure, our method, as with other state-of-the-art methods, is able to remove mitochondria (appear as a dark shaded texture) as well as vesicles (appear as small bubbles). An uncertain case is illustrated by a gray region, where the proposed network must decide whether the highlighted pixels should be segmented as membrane. This region is ambiguous because of smearing of the membrane due to anisotropy in the data. Our approach outperformed other state-of-the-art methods on several standard metrics, such as foreground-restricted Rand scoring after border thinning ($V_{rand}$) and foreground-restricted information theoretic scoring after border thinning ($V_{info}$). A more detailed description of those metrics is available [2]. Quantitative results are summarized in Table 2. If only four levels of resolutions (i.e., three downsampling and upscaling blocks - submission on May 2016) are used, we can achieve better results in comparison with U-net [25]. On the other hand, when a full five-level network (or i.e., four downsampling and upscaling blocks - submission on Nov 2016) is deployed on the test data, our $V_{rand}$, $V_{info}$ scores are higher than those of the network in network approach [22], the fused-architecture approach [8] and the long-short term memory (LSTM) approach [30]. These assessments faithfully confirmed that a deeper architecture with a residual bottleneck helps to increase the accuracy of the EM segmentation task. In case mild post-processing is applied (we perform a 2D median filter with $radius = 2px$ for each slice of prediction), we are in the top three ranking, while the first-place method used a compute-intensive, complicated lifting multi-cut algorithm [3] to further enhance the result from the network. Developing (or applying) such advanced post-processing methods is left for the future work.

### 4.2. Larval Zebrafish EM data

The larval zebrafish serial-section EM data analyzed here was captured from a 5.5 days post-fertilization larval zebrafish. This specimen was cut into 18000 serial sections and collected onto a tape substrate with an ATUM [13]. A series of images spanning the anterior quarter of the larval zebrafish was next acquired at a nearly isotropic resolution of $56.4 \times 56.4 \times \sim 60$ nm$^3$ vx$^{-1}$ from 16000 sections in the resulting serial section library using scanning EM. All image planes were then co-registered into a three-dimensional volume with an FFT signal whitening approach [33].

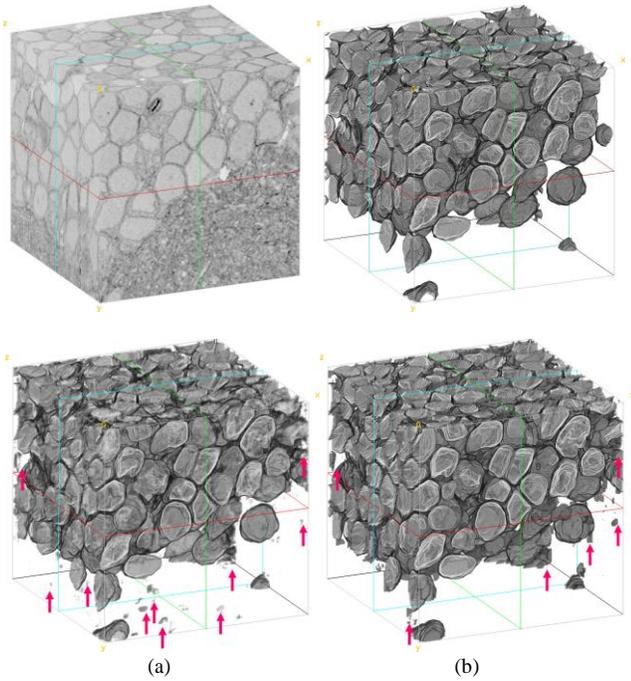

Figure 8. Volume rendering of the raw testset (a); its manual segmentation (b); U-net [25] prediction (c) and our prediction (d)

For training, two small sub-regions dataset crops were extracted from a near-final iteration of the full volume alignment in order to avoid ever deploying the segmentation on the training data. Two volumes were chosen to train on different features in the dataset. One volume was 512×512×512 and the other was 512×512×256. The features of interest, neuronal nuclei, were manually segmented as area-lists in each training volume using the Fiji [27] with TrakEM2 plug-in [7]. These area-lists were exported as bi- nary masks used in the training procedure. For accuracy assessments, an additional 512×512×512 sub-volume was manually segmented (test dataset). Figure 7 showed the manual segmentation of slice 195/512 in the test dataset, U-net prediction along side with the plain outcome from our network. In addition, Figure 8 displays the volume rendering of the test set in term of EM data, manual segmentation, U-net, and our method. As shown in these figures, the proposed architecture introduced less false prediction compared to U-net (indicated by red arrows). Table 3 compares U-net and our method using three quality metrics, e.g., $V_{rand}$, $V_{info}$ and the Dice coefficient $V_{dice}$, which also confirms that our method generates more accurate results than U-net.

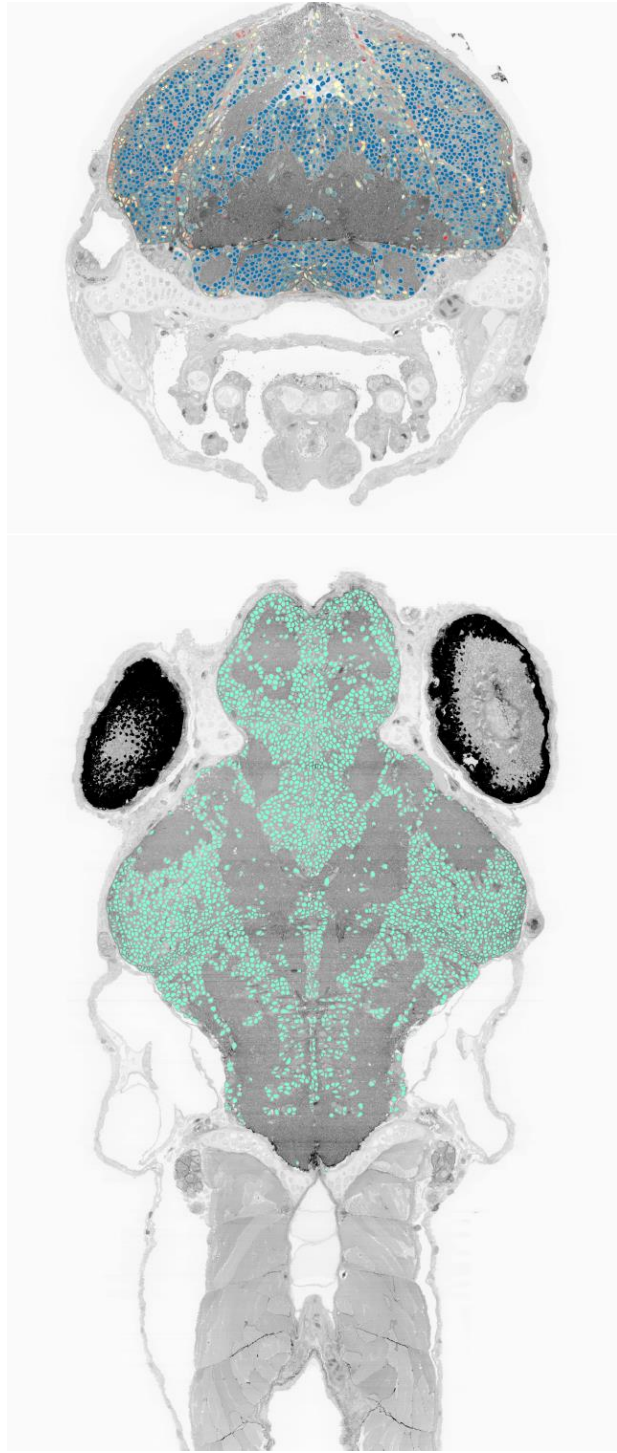

Figure 9. Overlaying the cell body segmentation on top of a larval zebrafish EM image: two viewpoints of the transverse (top, blue to red color map for cell sphericity) and horizontal (bottom) planes.

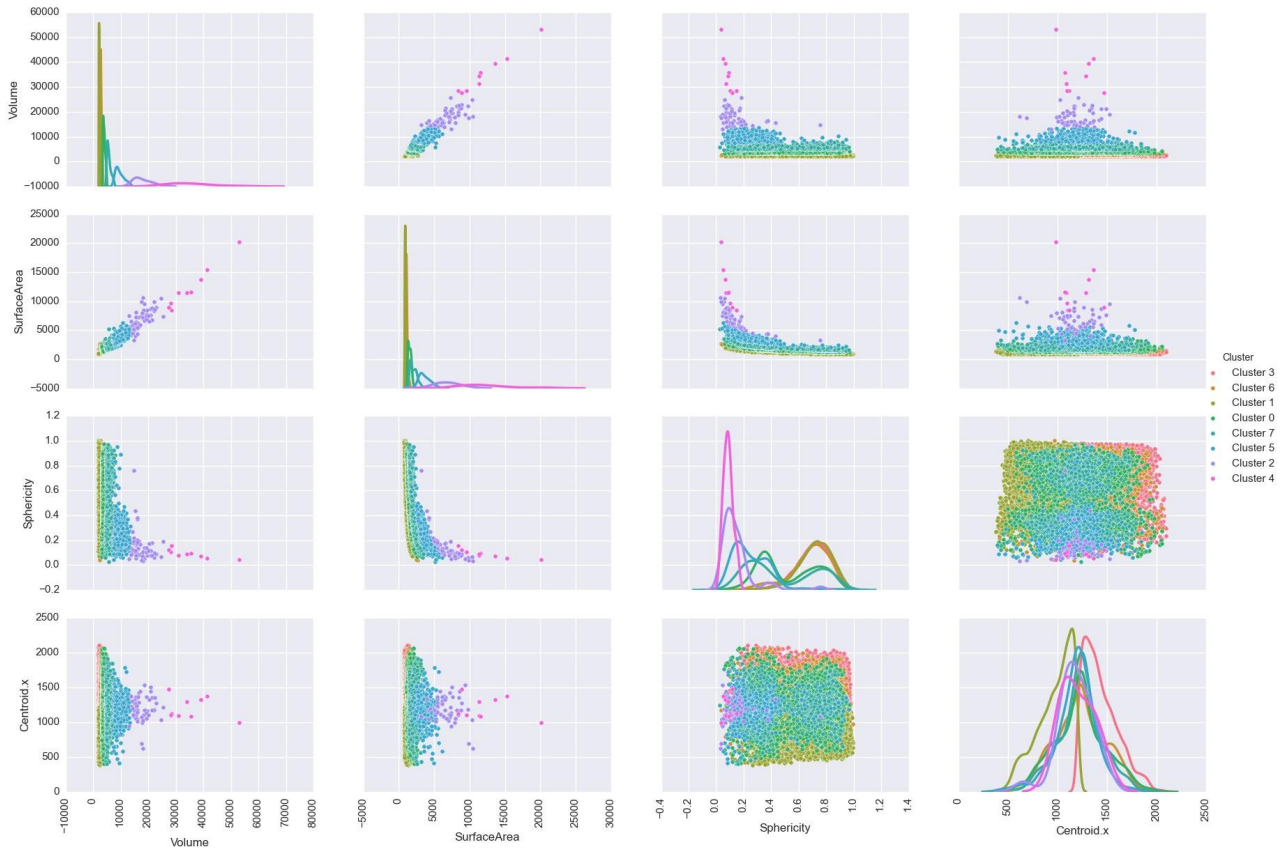

Figure 10. Correlation of statistical features: volume, surface area and sphericity, with horizontal coordinates of the centroids.

We deployed the trained network to the complete set of 16000 sections of the larval zebrafish brain imaged at 56.4×56.4× ∼60 nm$^3$ vx$^{-1}$ resolution, which is around 1.2 terabytes in raw data size. We discovered that there are approximately 180,000 cell bodies in the larval zebrafish brain. Figure 9 shows cross-sectional views (top: transverse (*x-y*) and bottom: horizontal (*x-z*)) of the EM volume overlaid with cell segmentation. The coronal view shows the sphericity (i.e., roundness) of the segmented cell body in a blue to red color map, which helps to identify the location of non-cell structures (i.e., false positive). We have conducted a statistical analysis of the cell morphological properties, such as volume, surface area, and sphericity. Those features are then clustered with a K-means algorithm and plotted in the form correlations, along side with their corresponding centroid coordinates. As illustrated in Figure 10, the outliers in the magenta cluster, correspond to cells with large volumes and surface areas, are clearly observed and can be eliminated by thresholding. We also observed that the feature distribution appears to be mirror symmetric across the midline of the larval zebrafish. This implies that the organization of neurons in the left and right brain hemisphere is likely to be very similar.

## 5. Conclusion

In this paper, we introduced a novel architecture of deep network for image segmentation that specifically targets connectomics image analysis. The proposed architecture, FusionNet, is a novel extension of U-net and residual CNN to develop a deeper network for a more accurate end-to-end connectomics image segmentation. We demonstrated the flexibility and performance of FusionNet in membrane- and blob-type EM segmentation tasks, and confirmed that FusionNet outperformed state-of-the-art methods in several standard quality metrics.

In the future, we plan to conduct in-depth analysis of FusionNet to get a better understanding of the architecture. We also plan to construct extremely deep FusionNet to improve segmentation accuracy. Developing distributed FusionNet for parallel training and deployment on a cluster system is another interesting research direction we wish to explore.